# CONTENT ADAPTIVE SCREEN IMAGE SCALING

*Yao Zhai*[1(*)], *Qifei Wang*[2], *Yan Lu*[2], *Shipeng Li*[2]

[1]University of Science and Technology of China, Hefei, Anhui, 230027, China
[2]Microsoft Research, Beijing, 100080, China

## ABSTRACT

This paper proposes an efficient content adaptive screen image scaling scheme for the real-time screen applications like remote desktop and screen sharing. In the proposed screen scaling scheme, a screen content classification step is first introduced to classify the screen image into text and pictorial regions. Afterward, we propose an adaptive shift linear interpolation algorithm to predict the new pixel values with the shift offset adapted to the content type of each pixel. The shift offset for each screen content type is offline optimized by minimizing the theoretical interpolation error based on the training samples respectively. The proposed content adaptive screen image scaling scheme can achieve good visual quality and also keep the low complexity for real-time applications.

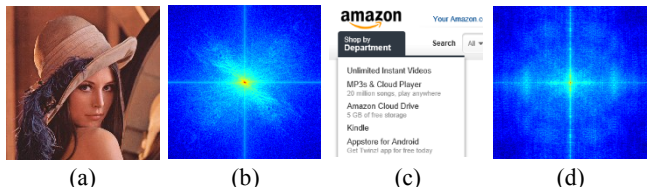

Fig. 1. (a) pictorial sample region; (b) DFT energy density of pictorial region; (c) text sample region; (d) DFT energy density of text region.

## 1. INTRODUCTION

Driven by the emerging applications like cloud computing, video conferencing, and desktop remoting, efficiently virtualize screen content on the screens with different resolutions becomes a hot research topic recently [1]. The resolution mismatch between the target screen and the original virtualized image may cause significant visual artifacts and severely degrade the user experience in screen applications. High efficiency screen image scaling is therefore highly desired by both the local and remote systems.

Although image scaling has been well studied for natural images [2-6], only a few literatures have been reported for screen image scaling. Essentially, the main challenge in image scaling is to *interpolate* millions of unknown pixel values based on the limited input sampled pixel values. However, different from the natural image that most pixels can be treated as isotropic units, the screen image virtualized by computer usually contains hybrid components like texts, pictures, and graphics, etc. All these components have quite different characteristics. As shown in Fig. 1, the energy of pictorial regions concentrates on the low frequency part in the discrete Fourier transform (DFT) domain. However, since text regions contain more sharp edges than pictorial regions, the energy density of text regions shows a different pattern from that of pictorial regions. Scaling screen image by the traditional uniform interpolation schemes, such as nearest neighbor, bilinear, and bicubic [2], usually suffers from the jaggy, over-smooth, and halo artifacts on sharp edges, respectively, which significantly degrade the visual quality.

Recently, many advanced adaptive image interpolation schemes were further proposed for natural image scaling, for example, the edge-based interpolation schemes [3][4] and the example-based interpolation schemes [5][6]. However, the high complexity makes them not practical for screen image scaling in real-time applications. Moreover, Sun et al. proposed a two-step screen image scaling algorithm which scales screen image via the bilinear interpolation first and further enhances the edges by adaptive sharpen filtering [7]. Although the content information has been exploited to assist adaptive sharpen filtering, this scheme also suffers from amplified noise in the sharp edge regions.

The screen images with good visual quality require clear texts and smooth pictures since these two parts attract most of the visual saliency. Therefore, efficient screen image scaling algorithm should be adaptive to both text and pictorial contents. To balance the visual quality and complexity, we propose a content adaptive screen image scaling scheme by combining the screen content analysis with a light-weight linear interpolation. Our scheme first classifies each region of the screen image into text or pictorial region according to the high gradient pixel number and the pixel color distribution. Next, each region is scaled by the shift linear interpolation (SLI) [8] with adaptive shift offsets. The adaptive shift offsets are offline optimized for the text and pictorial regions, respectively. The screen content analysis and the optimized adaptive shift offsets can finally achieve enhanced visual

---

quality in screen image scaling. Moreover, due to the linear interpolation kernel, the proposed scaling scheme retains low complexity so that it can be applied in real-time applications.

The reminder of this paper is organized as follows. Section II introduces the screen content classification. The proposed content adaptive screen image scaling scheme is described in Section III. The simulation results are shown in Section IV, followed by the conclusions in Section V.

## 2. SCREEN CONTENT CLASSIFICATION

To efficiently distinguish text and pictorial regions, two features are selected for classification. First, text regions usually contain more high gradient pixels than pictorial regions where the pixel values are smoothly changed. Second, other than the pictorial regions with widely spread pixel color distribution, the pixel colors of text regions usually concentrate on several base colors [9]. Based on these observations, a two-step screen content classification algorithm can thus be designed as follows.

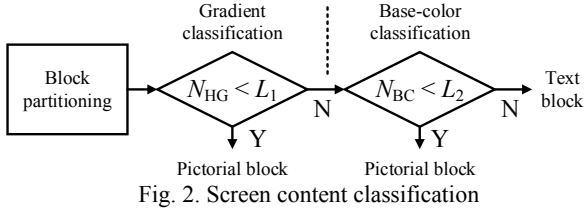

Fig. 2. Screen content classification

Before classification, each input screen image is partitioned into multiple 16x16 non-overlapped blocks. In the first step, the blocks with the number of high gradient pixels, $N_{HG}$, less than a pre-defined threshold $L_1$ will be classified as the pictorial blocks. The rest blocks are further classified by the base color criterion. The base color for each block is defined as the most frequent pixel color within the block. For each block, the number of base color pixels is further defined as the number of pixels with their color values falling within the range that is ±2 from the base color values. The blocks will be further classified as pictorial blocks if the number of base color pixels, $N_{BC}$, is larger than an adaptive threshold $L_2$. Here, $L_2$ is adjusted to a higher value if its left, upper and upper-left blocks are all pictorial blocks. Otherwise, it is set to a lower value. Both the higher and lower thresholds for $L_2$ are offline determined with the training samples. Fig. 2 demonstrates the content classification process.

After the content type is determined for each block, all the pixels within one block are assigned to the type of the block they belong to. The content type of each pixel will further determine its weights in interpolation.

## 3. CONTENT ADAPTIVE SCREEN IMAGE SCALING

To achieve good visual quality of screen image scaling with low complexity, we adopt the SLI scheme to predict the new pixel values and further make that interpolation scheme adapted to different screen content types obtained by the content classification. In this section, the SLI scheme is first briefly introduced. The shift offset optimization and the proposed scaling scheme are further presented.

### 3.1. Shift linear interpolation

In the SLI scheme, the sub-pixels between two contiguous sampled pixels are interpolated by the piece-wise linear functions. However, different from the traditional bilinear scheme where a single linear function is applied between each two contiguous sampled pixels with the start and end points exactly locating on the sampled pixels, the SLI scheme introduces a shift offset $\tau$ to the start and end points of the piece-wise linear interpolation function as shown in Fig. 3. Consequently, the signal wave between two contiguous sampled pixels is approximated by two piece-wise linear functions. The $\tau$-shift linear interpolation can be expressed as
$$f_{int}(x) = \sum_{n \in \mathbb{Z}} c_n \Lambda(x - n - \tau), \quad (1)$$
with the coefficient $c_n$ being inducted by
$$c_n = -\frac{\tau}{1-\tau} c_{n-1} + \frac{1}{1-\tau} f_n. \quad (2)$$
Here, The interpolation kernel function $\Lambda(x)$ is a linear B-spline: $\Lambda(x) = 1 - |x|$ for $|x| \leq 1$ and $\Lambda(x) = 0$ for $|x| > 1$. $f_n$ denotes the value of the sampled pixel $n$. The SLI scheme can thus be treated as a two-step process that Eq. (2) represents the first pre-filtering step mapping $f_n$ to $c_n$ and Eq. (1) represents the second interpolation step.

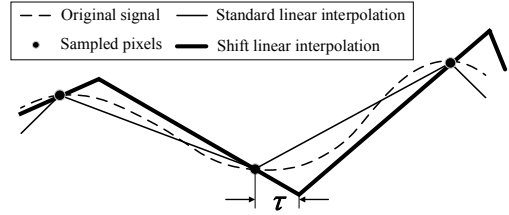

Fig. 3. Shift linear interpolation

### 3.2. Shift offset optimization for screen contents

The separable interpolation error of 2D screen image can be estimated in the frequency domain by the following function:
$$\eta(T) = \sqrt{\frac{1}{2\pi} \sum_{\omega_h} \sum_{\omega_v} |F(\omega_h, \omega_v)|^2 E(\omega_h T) E(\omega_v T)}, \quad (3)$$
where $\omega_h$ and $\omega_v$ denote the horizontal and vertical angular frequencies, respectively, $T$ denotes the pixel interval, $|F(\omega_h, \omega_v)|^2$ denotes the energy density of the input image, $E(\omega_h T)$ and $E(\omega_v T)$ are what we call the Fourier error kernels for horizontal and vertical interpolation, respectively. For the SLI scheme defined by Eqs. (1) and (2), the Fourier error kernel $E_\tau(\omega)$ can be expressed as:
$$E_\tau(\omega) = 1 + \frac{1}{3} \frac{2+\cos\omega}{|1-\tau+\tau e^{-j\omega}|^2} - 2\text{sinc}^2\left(\frac{\omega}{2\pi}\right) \text{Re}\left(\frac{e^{-j\omega\tau}}{1-\tau+\tau e^{-j\omega}}\right).$$
(4)

Since the interpolation error in Eq. (3) highly depends on the signal energy density, and the Fourier error kernel of SLI is related to the shift offset $\tau$, the performance of SLI can be optimized by adapting $\tau$ to minimize the interpolation error for different screen content, respectively.

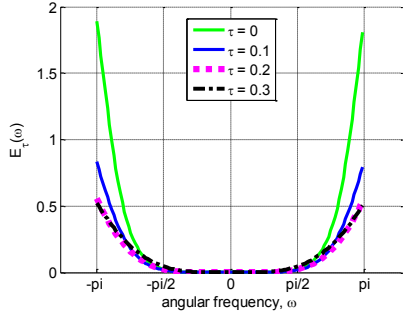

Fig. 4. Interpolation error kernel curves with different shift offset

To optimize the shift offset $\tau$, we have to obtain the Fourier error kernels and the energy densities for both the text and pictorial regions. Fig. 4 illustrates the curves of $E_\tau(\omega)$ against the shift offset $\tau$ that larger $\tau$ results in less error for high frequency components. Additionally, the energy densities of text and pictorial regions can be obtained based on some sample text and pictorial regions which are extracted from the training screen images. In this paper, we captured 200 16x16 sample text blocks and 200 16x16 sample pictorial blocks to train $\tau$ for text and pictorial regions, respectively. The average vertical energy densities of the sample text and pictorial blocks are shown in Figs. 5 (a) and (b), respectively. Subsequently, the vertical interpolation errors of the text and pictorial blocks which are obtained by Eq. (3) with varying $\tau$ are shown in Figs. 6 (a) and (b), respectively. Obviously, the interpolation error curves of both text and pictorial blocks are convex and the minimal distortion can be reached with a certain $\tau$ for each. In this paper, we use a polynomial function to fit the interpolation error against the shift offset $\tau$.

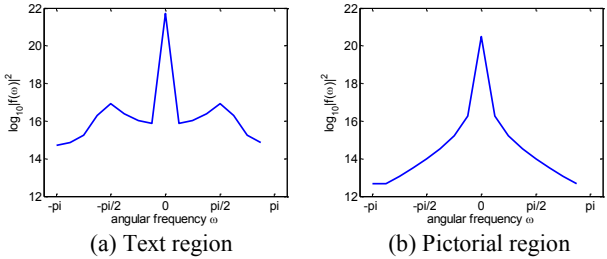

(a) Text region     (b) Pictorial region

Fig. 5. Energy densities of text and pictorial regions

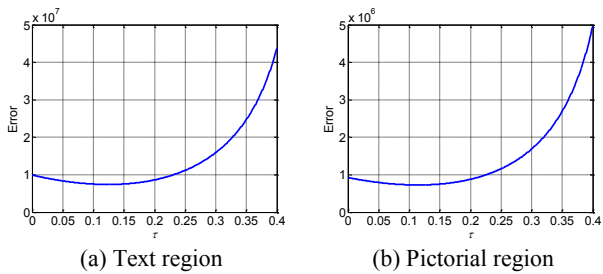

(a) Text region     (b) Pictorial region

Fig. 6. Error of SLI vs. different shift offset

Consequently, the optimized shift offset $\tau$ for vertical interpolation are 0.124 and 0.114 for text and pictorial regions, respectively. Similarly, the optimized offset $\tau$ for horizontal interpolation can be obtained by the same way. Finally, the optimized $\tau$ is summarized by Table I.

Table I Optimized $\tau$ for text and pictorial regions

|  | Text region | Pictorial region |
|---|---|---|
| Horizontal | 0.110 | 0.112 |
| Vertical | 0.124 | 0.114 |

In the proposed scaling scheme, the shift offset $\tau$ in the first pre-filtering step is adaptively adjusted according to the content type of each pixel. To avoid the anisotropic pixel phase shift in SLI and the blocking artifacts caused by the screen content classification, the shift offset $\tau$ in the second interpolation step should be operated with a global value. In our algorithm, the global shift offset is set according to the major pixel type which is the most frequent block type in the current frame.

### 3.3. Algorithm implementation and complexity analysis

Based on the optimized $\tau$ for text and pictorial regions, the content adaptive screen image scaling algorithm can be implemented in a separable mode by the following steps:
1) Determine the content type for each pixel in the current picture by the content classification algorithm. Further determine the major pixel type for the current picture.
2) Pre-filter each sampled pixels in horizontal direction based on the content type of each pixel by Eq. (2).
3) Interpolate the new pixels horizontally by Eq. (1). The interpolation shift offset is set by the major pixel type of the current image.
4) Pre-filter the interpolated pixel values from Step 3 vertically by Eq. (2). The content type of each interpolated pixel is set according to its nearest neighboring sampled pixel.
5) Interpolate the new pixels vertically by Eq. (1) with the interpolation shift offset obtained in Step 3.

Finally, the output image is scaled to the desired resolution.

To evaluate the complexity of our proposed screen image scaling scheme, we compare the operation number of the proposed scheme with that of other traditional interpolation schemes. In SLI, the first pre-filtering step requires $4N_{in}$ multiplication and addition operations, and the second interpolation step requires $8N_{out}$ multiplication and addition operations. $N_{in}$ and $N_{out}$ denote the pixel numbers of input and output images, respectively. Beyond SLI, the proposed content adaptive scaling scheme involves an additional content classification step. Assume all the blocks go through the two steps in content classification, it will cost $4N_{in}$ additions for the first step and $N_{in} + 4N_b$ additions for the second step. $N_b$ denotes the block number in the original image. Therefore, the overall operations of the proposed scaling scheme should be less than $9N_{in} + 8N_{out} + 4N_b$ addition plus $4N_{in} + 8N_{out}$ multiplication. Compared with the bilinear scheme which requires $8N_{out}$ multiplication and addition operations and the bicubic scheme which requires $44N_{out}$ multiplication and addition operations, the proposed scaling scheme retains the low complexity and can be applied in the real-time applications.

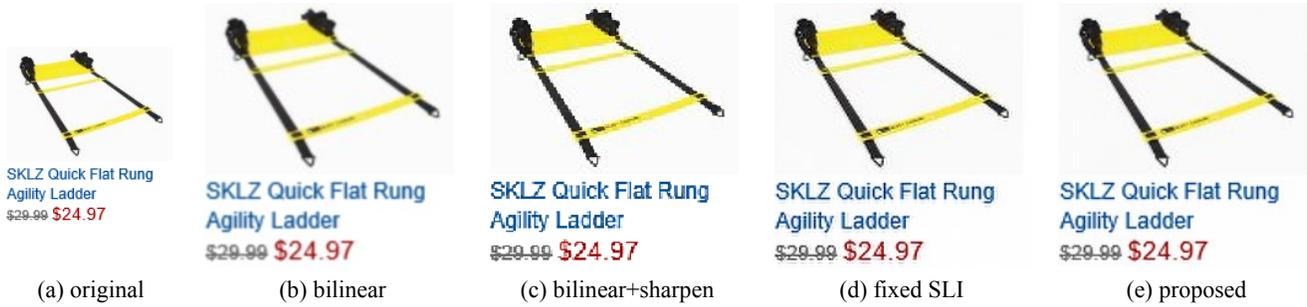

(a) original     (b) bilinear     (c) bilinear+sharpen     (d) fixed SLI     (e) proposed

Fig. 7. Original screen image scaling results

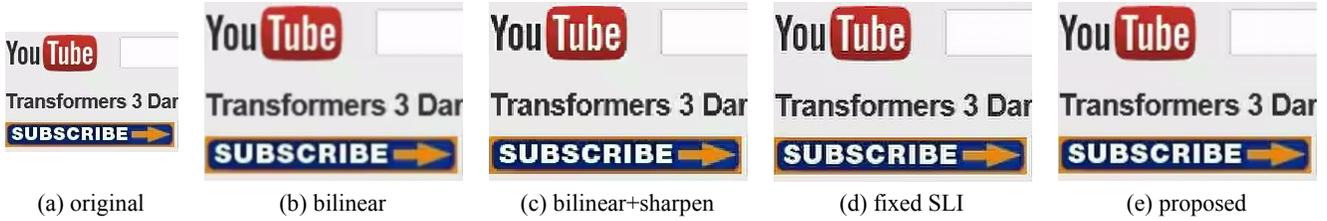

(a) original     (b) bilinear     (c) bilinear+sharpen     (d) fixed SLI     (e) proposed

Fig. 8. Decoded screen image scaling results

## 4. SIMULATION RESULTS

In this section, the simulation results are demonstrated to evaluate the performance of the proposed content adaptive screen image scaling algorithm. The simulation results of some traditional algorithms, including bilinear scheme, bilinear plus adaptive sharpen filtering scheme [7], and the SLI scheme with the shift offset fixed to the general asymptotic optimal value, 0.21 [8], are also shown in this section for visual quality comparison.

The scaling simulations are carried out on both the original captured test screen images for local applications and the decoded images of the original captured ones for remote applications. The coding simulations are implemented by x264 codec [10] with the encoding QP = 37. The resolution of the input screen images is 1280x768. All the images are up-scaled by the factor of 1.5. The output resolution is thus 1920x1152. The test screen images contain hybrid contents including texts and pictures.

Fig. 7 illustrates the original image scaling results of the four test schemes. For simplicity, we use "original", "bilinear", "bilinear+sharpen", "fixed SLI", and "proposed" to denote the original reference image, the scaling results generated by bilinear, bilinear plus adaptive sharpen filtering, SLI scheme with fixed shift offset, and our proposed scheme, respectively. It can be seen from Fig. 7 that the bilinear scheme seriously blurs both text and pictorial regions. The bilinear plus sharpen filtering scheme introduces strong jaggy artifacts on the edges in both text and pictorial regions which also significantly degrade the visual quality. Since the global asymptotic optimal shift is generally derived, the fixed SLI scheme over sharpens some input pixel values in the pre-filtering step and results in the ghost artifacts in text regions. Finally, our proposed scheme achieves both clear texts and smooth pictures because the shift offset is adaptively adjusted to the optimized value according to the content type.

Fig. 8 demonstrates the decoded image scaling results of the four test schemes. The "original" label here denotes the decoded image before scaling. Similarly, the bilinear scheme also seriously blurs the texts and pictures. Besides the jaggy artifacts in the bilinear plus sharpen filtering scheme and the ghost artifacts around the text region in the fixed SLI scheme, these two schemes both augment the coding distortion due to the over sharpening. Since our proposed scheme balances the sharpening and low-pass filtering with the adaptive shift offset, it finally achieves clear texts and smooth pictures with reduced noise for the decoded screen images.

In the simulations, the proposed scheme increases the processing time of bilinear scheme by 30% and reduces the processing time of bicubic scheme by 75% on average. Compared with the fixed SLI scheme, the average processing time rise of the proposed scheme is only 8%.

In summary, our proposed content adaptive scaling scheme obtains balanced performance of visual quality and complexity for both the original and decoded screen images.

## 5. CONCLUSIONS

This paper proposes a content adaptive screen image scaling algorithm by combining the screen content analysis and the adaptive shift linear interpolation (SLI). The content analysis algorithm first classifies input screen images into text and pictorial regions. The subsequent adaptive SLI predicts the new pixel values with the content adaptive shift offsets. The shift offset is offline optimized for each screen content type, including texts and pictures, by minimizing the theoretical interpolation error with some training samples respectively. Based on the simulation results, the proposed scheme can generate good visual quality, for example, clear texts, smooth pictures, and also reduced noise, for both the original and decoded screen images. It also maintains the low complexity desired by the real-time applications.